\documentclass[10pt,a4paper,onecolumn]{article}

\usepackage[utf8]{inputenc}
\usepackage{amsmath}
\usepackage{amsfonts}
\usepackage{amssymb}
\usepackage{hyperref}
\usepackage{graphicx}

\newif\ifdraft
\draftfalse

\newcommand{\swn}{\textsc{SentiWordNet}}
\newcommand{\wn}{\textsc{WordNet}}

\title{The User Feedback on \swn}

\author{Andrea Esuli\\
Istituto di Scienza e Tecnologie dell'Informazione ``A. Faedo''\\
via G. Moruzzi, 1 -- 56124 Pisa, Italy\\
\nolinkurl{andrea.esuli@isti.cnr.it}}

\date{June 4, 2013}

\begin{document}

\maketitle


\begin{abstract}
With the release of \swn\ 3.0 the related Web interface has been restyled and improved in order to allow users to submit feedback on the \swn\ entries, in the form of the suggestion of alternative triplets of values for an entry.
This paper reports on the release of the user feedback collected so far and on the plans for the future.
\end{abstract}


\section{Introduction}

\swn\ \cite{Baccianella2010} is an automatically generated lexical resource that assigns to each sysnset of \wn\ \cite{Miller1995} a triplet of positivity, negativity and objectivity scores.

\swn\ is a \emph{general} (or \emph{global}) lexicon, i.e., its scores are deemed to be of general application regardless of the specific domain of the text which contains the terms to which the scores are associated.
The hypothesis on global applicability of scores to terms may not hold on all the possible uses of terms in any domain but, from a practical perspective, it should hold for a large part of the cases.
In \swn\ this hypothesis holds stronger because the application of scores to each distinct sense of a term allows to discriminate a good number of otherwise sentiment-ambiguous cases, e.g., \texttt{blue} in the sense of the color or in the sense of the mood\footnote{This also requires sense level disambiguation of text.}.

In its practical use \swn\ can be considered a resource that provides a basic, wide-coverage, sentiment knowledge into the application in which it is used.
Domain-specific applications will then likely pair it with a domain-specific resource in order to improve the precision on domain-specific terms.

\begin{figure}[htbp]
\centering
\includegraphics[width=\textwidth]{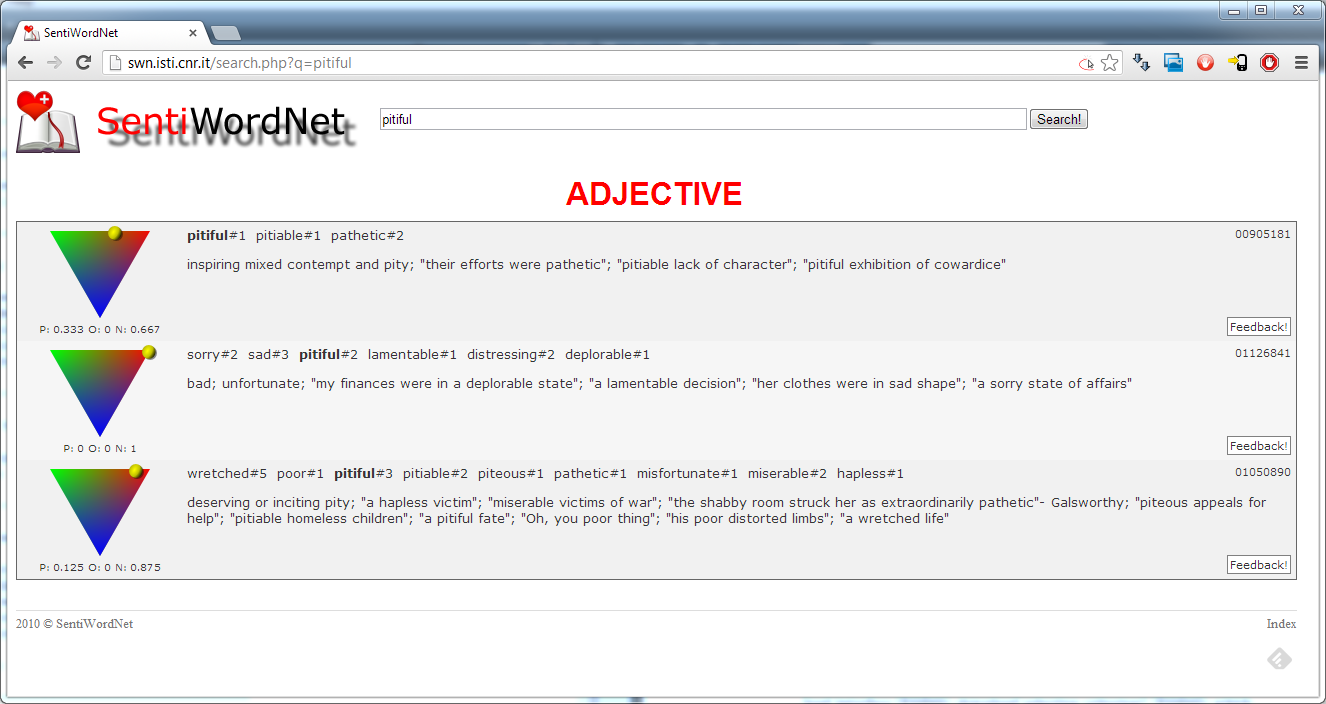}
\caption{\label{fig:browse}The \swn\ web interface with "Feedback!" button.}
\end{figure}

\swn\ is an automatically generated resource and, as for any other automatic labeling process based on machine learning, it contains errors, i.e., incorrect triplets of values associated to some synsets.
With the release of \swn\ 3.0 the related Web interface\footnote{\url{http://swn.isti.cnr.it/}} has been restyled and improved in order to allow users to submit feedback on the \swn\ entries, in the form of the suggestion of alternative triplets of values for an entry, as shown in Figures \ref{fig:browse} and \ref{fig:oldfeed}.

In the following we report on the collected feedback information and also on the future evolution of the feedback collection process.


\section{Collecting the feedback}

\begin{figure}[htbp]
\centering
\includegraphics[width=\textwidth]{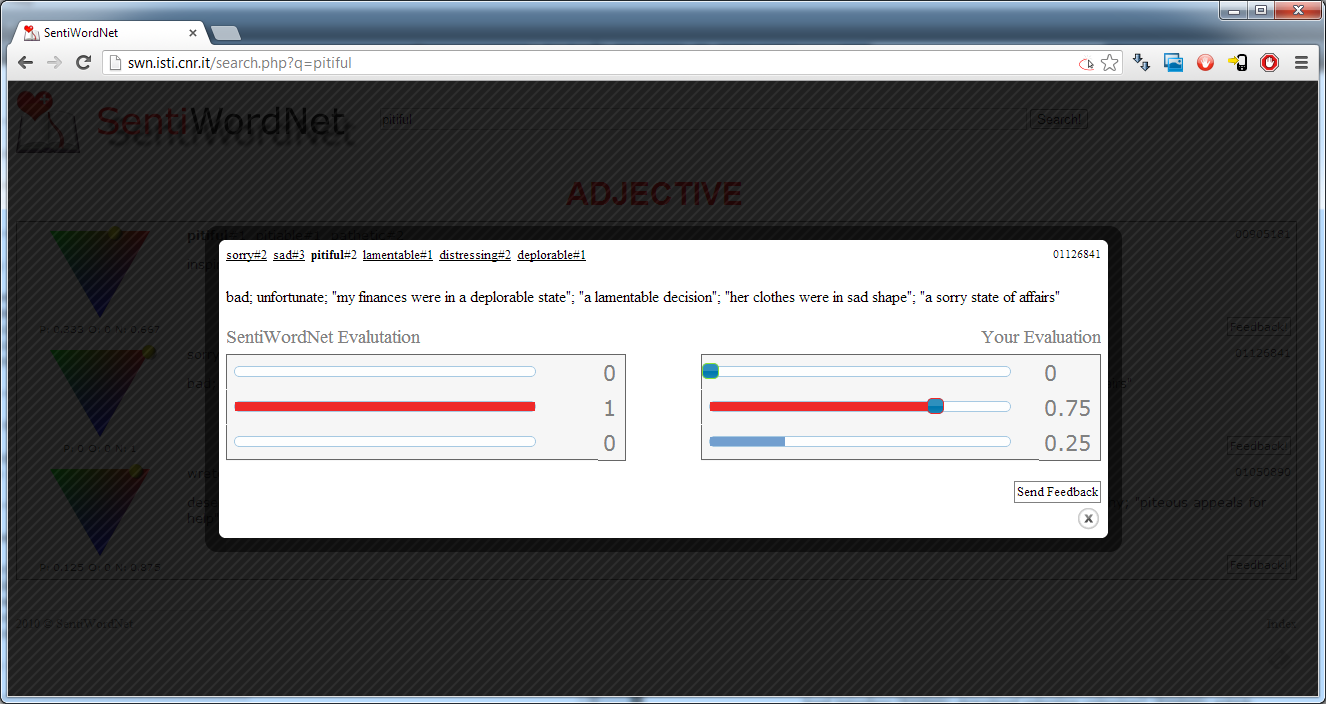}
\caption{\label{fig:oldfeed}The feedback dialog.}
\end{figure}

\begin{figure}[htbp]
\centering
\includegraphics[width=\textwidth]{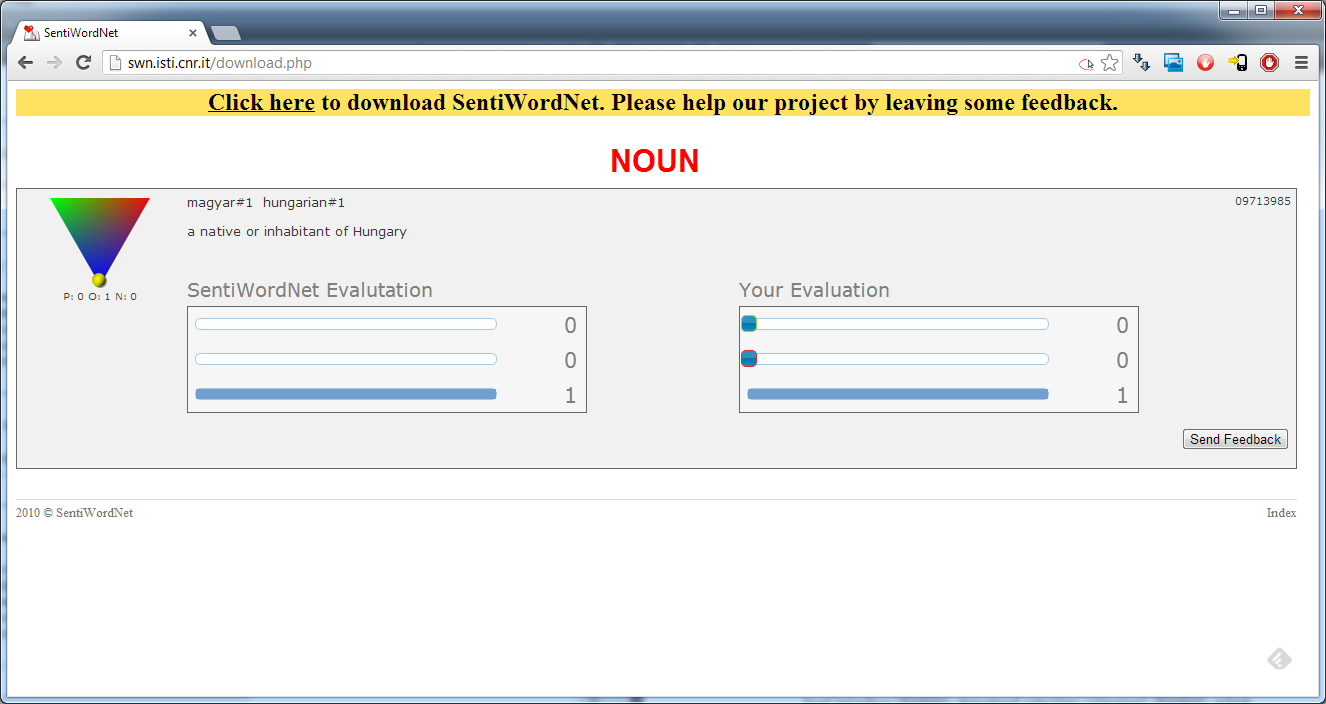}
\caption{\label{fig:down}The feedback dialog presented when \swn\ is downloaded.}
\end{figure}

Feedback from users has been collected in two ways: (i) from the Web interface, and (ii) from the download interface.

The Web interface allows the user to search for any term in \wn\ and shows all the synsets to which the term belong with the relative \swn\ values.
Each synsets has a ``Feedback!'' link that opens a feedback dialog (see Figure \ref{fig:oldfeed}) that allows the user to submit its suggested triplet of values (which can differ or be the same of the \swn\ triplet).

When a user clicks on the link to the download page of \swn\ the actual download link is presented along with a feedback dialog (see Figure \ref{fig:down}) that presents the triplet for a synset and asks to the user to check them, confirm it or submit a triplet of corrected values.
After the submission of the feedback the interface loads automatically the values for another synset, so that the user can submit feedback on other \swn\ triplets.
Synsets are selected in random order among the set of synsets with the minimum number of feedbacks.
This means, for example, that whenever a synset gets its first feedback, independently of the feedback source (web search, or download link), it will be never presented for feedback from the download link until \emph{all} synsets will have at least one feedback too.

When users submit their feedback the information that is stored is:
\begin{itemize}
\item Original \swn\ values (positivity and negativity, the objectivity value can be determined as $1-(\mbox{posivitity}+\mbox{negativity})$).
\item The values submitted by the user (positivity and negativity).
\item IP of the browser submitting the data.
\item Time and date.
\end{itemize}


\section{Releasing the feedback}

Before releasing the feedback we faced with the requirement of preserving the privacy of the users that have contributed their feedback.
At the same time we liked to save some information about the event that generated the feedback, which could be useful for the analysis of data.

We thus converted each IP to a unique identifier, so that it is still possible to group feedbacks generated by the same IP but not to determine its exact source.
We also added the country in which the IP is located, by using the \emph{WHOIS} protocol.
Though the information obtained in this way is only a rough approximation of the location of the feedback source, and it does not provide any information about the fluency in English of the submitter, it can be useful for future analyses involving grouping or filtering based on geographic areas.

The exact time information has been removed, leaving only the date of submission.

The feedback entries are released in a text file, with a feedback entry per line, and each line composed of the following tab-separated field:

\begin{itemize}
\item Incremental counter: can be used as unique identifier of feedback entries.
\item SynsetId: the offset in the \wn\ data file. Along with the part of speech it uniquely identifies the synset.
\item Part of speech of the synset.
\item \swn -positivity: the positivity value assigned to the synset in \swn\ at the time of feedback submission.
\item \swn -negativity: same as above, for negativity.
\item Feedback-positivity: the positivity value submitted by the user.
\item Feedback-negativity: same as above, for negativity.
\item Date of submission.
\item Anonymized IP.
\item Country in which the IP is locate: two letter country code, ``xx'' if not available.
\item List of word\#senseNumber pairs belonging to the synsets: attached to allow quick human inspection without need to search for the synset on \wn .
\end{itemize}

The feedback data can be downloaded from the \swn\ website.
The feedback data is released under the \emph{Creative Commons Attribution-ShareAlike 3.0 Unported (CC BY-SA 3.0)} license\footnote{\url{http://creativecommons.org/licenses/by-sa/3.0/}}.


\section{User feedback statistics}

A total of 3510 feedback entries have been collected (about 3.5 a day on average).

The feedback entries have been submitted from 1209 distinct IP addresses. 
The average is thus 2.9 entries per IP address, and the distribution follows a power law, as shown in Table \ref{tab:ipdistr}.

\begin{table}[htbp]
\centering
\begin{tabular}{|r|r|}\hline 
 & \multicolumn{1}{c|}{Number of IP addresses}\\
\multicolumn{1}{|c|}{$f$} & \multicolumn{1}{c|}{with at least}\\
 & \multicolumn{1}{c|}{$f$ feedback entries}\\ 
\hline 
1 & 1209 \\  
2 & 522 \\ 
5 & 153 \\
10 & 47 \\  
20 & 12 \\  
50 & 4 \\  
100 & 1 \\  
200 & 1 \\ 
\hline 
\end{tabular} 
\caption{\label{tab:ipdistr} Cumulative distribution of IP addresses with respect to the number of feedback entries submitted.}
\end{table}

With respect to the geographic origin of the feedback, 350 of the 1209 distinct IP addresses cannot be mapped to any country (``xx'', in Table \ref{tab:country}). 
The remaining IP addresses come from 72 countries, 35 of which have contributed at least 15 feedback submissions from at least 5 distinct IP addresses.

\begin{table}[htbp]
  \centering
  \resizebox{\textwidth}{!}{
    \begin{tabular}{|c|r|r||c|r|r||c|r|r|}
    \hline
    Country & \#IPs & \#subs & Country & \#IPs & \#subs & Country & \#IPs & \#subs \\
    \hline
    xx    & 350   & 1063  & SE    & 9     & 41    & MA    & 2     & 12 \\
    IN    & 145   & 292   & PH    & 9     & 21    & BG    & 2     & 9 \\
    GB    & 59    & 189   & IE    & 9     & 15    & CU    & 2     & 9 \\
    DE    & 59    & 115   & HK    & 9     & 14    & JO    & 2     & 9 \\
    IT    & 54    & 324   & RU    & 8     & 23    & LB    & 2     & 4 \\
    CN    & 50    & 157   & CH    & 7     & 35    & MT    & 2     & 3 \\
    ES    & 32    & 64    & NO    & 7     & 29    & LK    & 2     & 2 \\
    FR    & 29    & 57    & TN    & 6     & 36    & NZ    & 2     & 2 \\
    NL    & 26    & 103   & TH    & 6     & 33    & QA    & 1     & 3 \\
    SG    & 24    & 40    & DK    & 6     & 18    & US    & 1     & 3 \\
    AU    & 22    & 55    & VN    & 6     & 14    & UY    & 1     & 3 \\
    BR    & 19    & 41    & SI    & 5     & 25    & KP    & 1     & 2 \\
    MX    & 17    & 33    & CO    & 5     & 15    & MO    & 1     & 2 \\
    RO    & 16    & 40    & DZ    & 5     & 13    & NP    & 1     & 2 \\
    PK    & 15    & 37    & CL    & 5     & 10    & VE    & 1     & 2 \\
    TW    & 15    & 28    & PL    & 5     & 8     & AR    & 1     & 1 \\
    KR    & 14    & 70    & HU    & 4     & 45    & BA    & 1     & 1 \\
    TR    & 14    & 25    & FI    & 4     & 8     & EC    & 1     & 1 \\
    ID    & 13    & 18    & CZ    & 4     & 7     & IQ    & 1     & 1 \\
    PT    & 12    & 85    & EG    & 4     & 7     & KE    & 1     & 1 \\
    JP    & 12    & 26    & IR    & 4     & 7     & LT    & 1     & 1 \\
    GR    & 11    & 29    & AT    & 4     & 6     & MK    & 1     & 1 \\
    MY    & 10    & 33    & UA    & 3     & 15    & SY    & 1     & 1 \\
    IL    & 10    & 28    & HR    & 3     & 9     &       &       &  \\
    BE    & 10    & 24    & EU    & 3     & 5     &       &       &  \\
    \hline
    \end{tabular}
    }
  \caption{\label{tab:country}Distribution of IP address and feedback submissions by country.}
\end{table}

The distribution in time of the feedback submissions (see Table \ref{tab:timedistr}) shows a first period, in 2010, with a relatively low number of submissions, a second period in 2011-2012 with a higher steady flow of submissions (85 per month on average) and a high peak in the first months of 2013.

\begin{table}[htbp]
  \centering
    \begin{tabular}{|c|r|r|r|r|}
    \hline
    month/year & 2010  & 2011  & 2012  & 2013 \\
    \hline
    Jan   &       & 64    & 29    & 110 \\
    Feb   &       & 84    & 138   & 530 \\
    Mar   &       & 82    & 100   & 333 \\
    Apr   &       & 36    & 137   & 270 \\
    May   &       & 94    & 138   & 95 \\
    Jun   & 39    & 93    & 79    &  \\
    Jul   & 10    & 49    & 52    &  \\
    Aug   & 6     & 111   & 86    &  \\
    Sep   & 8     & 57    & 101   &  \\
    Oct   & 14    & 146   & 82    &  \\
    Nov   & 10    & 61    & 109   &  \\
    Dec   & 37    & 45    & 75    &  \\
    \hline
    \end{tabular}
  \caption{\label{tab:timedistr}Distribution in time of feedback submissions.}
\end{table}

Among all the feedback entries 2236 (63.7\%) have the same exact values of the original \swn\ triplet.
This in principle means that that feedback states the correctness of the relative \swn\ values, but there is also the possibility that part of that feedback has been generated by careless clicks on the feedback submission interface.
The evaluation of feedback quality is beyond the scope of this report, though we are working on a number of improvements on the feedback collection process aimed at improving the quality of the collected feedback, as described in the next section.


\section{Improving user feedback collection}

\begin{figure}[htbp]
\centering
\includegraphics[width=\textwidth]{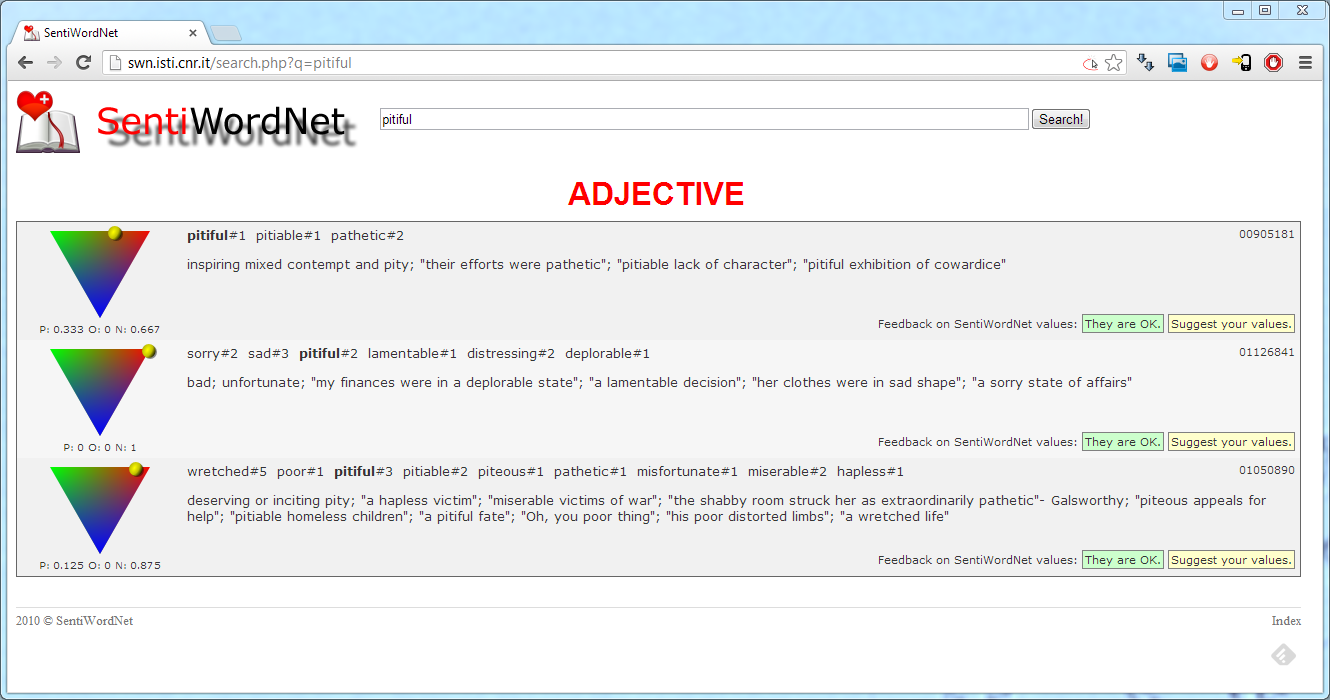}
\caption{\label{fig:newfeed}The new, more informative, feedback area in the \swn\ web interface.}
\end{figure}

While preparing feedback data we have noticed some weak points in the feedback collection process and we have designed some modification to the interface and the collected information in order to improve them.

The feedback submission interface presented from the download link is vulnerable to repeated submission of feedback that confirms the original \swn\ values.
This is due to the fact that that kind of submission has almost no ``cost'', i.e., it requires only to click the ``Submit feedback'' button, and that repeated click of that button does not even require to move the mouse.
In order to limit the false confirmatory submissions we have added a confirmation dialog that asks to the user to actively confirm the submission of the feedback.

We noted also that there is a difference in the users' role between the two possible sources of feedback, i.e., the download link and the search interface.
The feedback interface on the download link pushes requests of feedback to the users, with the synsets been evaluated selected by the interface, not by the users.
Users may be thus not motivated to give their evaluation or also may not be prepared to evaluate a specific synset (e.g., being non-native speakers of the language).
When users instead submit their feedback from the search interface, they do it on entries they have searched for.
Moreover, the main purpose of the interface is not feedback submission and thus it is likely that when the user decides to submit feedback the submitted values are carefully thought.

In the original implementation of the feedback collection process we did not differentiate the sources of feedback.
It is not possible, on the feedback data from this first release, to tell if the feedback comes from the download or the search interface.
We have added this information to any feedback collected from now on.

We have also redesigned the original feedback area in the search interface, substituting the single ``Feedback!'' button (see Figure \ref{fig:browse}) with two buttons: one that explicitly confirms the correctness of the presented values and one that opens the dialog for the submission of an alternative triplet of values (see Figure \ref{fig:newfeed}).


\bibliographystyle{abbrv}
\bibliography{SWNFeedback}

\end{document}